\documentclass[journal]{journal}
\usepackage{mathtools}
\usepackage{amsmath}

\usepackage{ifpdf}

\usepackage{graphicx}
\usepackage{amsmath}
\usepackage{amsfonts}
\usepackage{color}
\usepackage{cite}
\usepackage{float}
\usepackage{listings}
\usepackage[utf8]{inputenc}
\graphicspath{ {figures/} }
\usepackage{array}
\usepackage{subcaption}
\usepackage{dsfont}  

\usepackage{booktabs,subcaption,amsfonts,dcolumn} 
\newcolumntype{d}[1]{D..{#1}} 
\usepackage{longtable,xtab,booktabs}

\usepackage{blindtext}
\usepackage[T1]{fontenc}
\usepackage{mathtools}
\usepackage[table,x11names]{xcolor}

\usepackage{hhline}

\usepackage{tikz} 
\usepackage{qtree}
\usepackage{tikz-qtree}

\allowdisplaybreaks
\usepackage{multirow}

\usepackage{cite}

\ifCLASSINFOpdf
\else
\fi
\usepackage{algorithmic}

\usepackage{array}

\usepackage{mdwmath}
\usepackage{mdwtab}

\usepackage{eqparbox}

\usepackage{stfloats}
\begin{document}
%
\title{The joint node degree distribution in the Erd\H{o}s-Rényi network}
%
%
%

\author{Boshra~Alarfaj,
        Charles~Taylor,
        and~Leonid~Bogachev
\thanks{B. Alarfaj is with the Department
of Statistics, University of Leeds, Leeds,
LS2 9JT UK, and Department of Statistics and Operations Research, College of Science, King Saud University, P.O.Box 2455, Riyadh 11451, SA, e-mail: balarfaj@ksu.edu.sa}
\thanks{C. Taylor and L. Bogachev are with the University of Leeds.}}

%
%

\markboth{Journal of \LaTeX\ Class Files,~Vol.~6, No.~1, January~2007}%
{Shell \MakeLowercase{\textit{et al.}}: Bare Demo of IEEEtran.cls for Journals}
%



\maketitle
\thispagestyle{empty}

\begin{abstract}
The Erd\H{o}s-Rényi random graph is the simplest model for node degree distribution, and it is one of the most widely studied. In this model, pairs of $n$ vertices are selected and connected uniformly at random with probability $p$, consequently, the degrees for a given vertex follow the binomial distribution. If the number of vertices is large, the binomial can be approximated by Normal using the Central Limit Theorem, which is often allowed when $\min (np, n(1-p)) > 5$. This is true for every node independently. However, due to the fact that the degrees of nodes in a graph are not independent, we aim in this paper to test whether the degrees of per node collectively in the Erd\H{o}s-Rényi graph have a multivariate normal distribution MVN. A chi square goodness of fit test for the hypothesis that binomial is a distribution for the whole set of nodes is rejected because of the dependence between degrees. Before testing MVN we show that the covariance and correlation between the degrees of any pair of nodes in the graph are $p(1-p)$ and $1/(n-1)$, respectively. We test MVN considering two assumptions: independent and dependent degrees, and we obtain our results based on the percentages of rejected statistics of chi square, the $p$-values of Anderson Darling test, and a CDF comparison.  We always achieve a good fit of multivariate normal distribution with large values of $n$ and $p$, and very poor fit when $n$ or $p$ are very small. The approximation seems valid when $np \geq 10$. 
We also compare the maximum likelihood estimate of $p$ in MVN distribution where we assume independence and dependence. The estimators are assessed using bias, variance and mean square error. 

\end{abstract}


%
\IEEEpeerreviewmaketitle

\section{Introduction}
\label{intro}
%
%
%
%
\IEEEPARstart{A}{random} graph is a set of nodes and edges in which (some or all) pairs of nodes are connected with edges at random. A simple example of a random graph is the Erd\H{o}s-Rényi \cite{ER1}, denoted by $G(n,p)$ which was first studied by Solomonoff and Rapoport \cite{Solomonoff}. In this model, there are $n$ nodes such that each pair of nodes is connected with an edge with independent probability $p$. Some properties of this model are mentioned in \cite{ER2} and \cite{ER3}. It is a simple graph which means there are no loops or multiple edges. It is an ensemble of networks, i.e. it is not generated as a single network, but in terms of a probability distribution over all possible graphs such that $P(G)=p^m (1-p)^{{n \choose 2} -m}$ for all simple graphs and zero otherwise, where $m$ is the number of edges in the graph $G$, \cite{Networks-Intro-2010}.
\

In the Erd\H{o}s-Rényi model, the probability that a node has a degree $x$ is Bin$(n-1,p)$ and the expectation and variance are $E(X)=(n-1) p$ and $Var(X)=(n-1) p (1-p)$, respectively.
For certain values of $n$ and $p$, we can approximate Binomial with Normal distribution, using the Central Limit Theorem, when $\min (np, n(1-p)) > 5$. We get a pretty good approximation when $p=0.5$ which means that the Binomial is perfectly symmetric, \cite{fischer2010history}. 

In many cases we are interested in the properties of large networks when the number of nodes is large. Hence, we can use the normal approximation to binomial independently for every single node in a graph. However, since the degrees of nodes are not independent, we suggest a multivariate normal distribution for the whole set of nodes collectively. In Section \ref{SS}, we use a $\chi^2$ goodness of fit to test the hypothesis that the binomial is a distribution for all nodes together. In Section \ref{corr}, we obtain the covariance and correlation between degrees. Then in Section \ref{MVN}, we conduct several tests in order to prove that MVN is a good distribution for all nodes with taking into account the dependence between degrees. We also compute the maximum likelihood estimate of the probability of an edge presence in a graph in Section \ref{MLE} assuming two cases: independent and dependent degrees. After that we make a comparison between the cases of independence and dependence which is in Section \ref{Comp}.

\section{Simulation to test if Binomial is the joint node degree distribution of the Erd\H{o}s-Rényi model}
\label{SS}

We test whether the degrees of each node collectively in the Erd\H{o}s-Rényi has Binomial distribution using $\chi^2$ goodness of fit test $\chi^2_{df}=\sum_{i=1}^n\left[ (O_i-E_i)^2/E_i\right]$
where $O_i$ and $E_i$ are the observed and expected frequencies of the degree $i$, respectively \cite{cochran1952chi}. The degrees of freedom is 
$df=\textrm{number of categories}-l -k$
where $l$ is the number of constraint, usually when the sum of expected counts adds up to the sum of observed counts, this is one constraint. $k$ is the number of estimated parameters \cite{df}. 10000 simulated graphs are generated and tested. All have the parameters $n=61$ and $p=0.1$. 
The expected counts of nodes $E_i$ are calculated by multiplying the Binomial probability by $61$. The small values are gathered into one category which are the values associating with the degrees (0 to 3) and those with (9 to 60). Therefore, the number of categories decreases from 61 to 7.
The observed counts of nodes are also summed up for the same categories of expectations. Consequently, the degrees of freedom is $df=6$. 

\begin{figure}[H]
\centering
\includegraphics[width=2.5in]{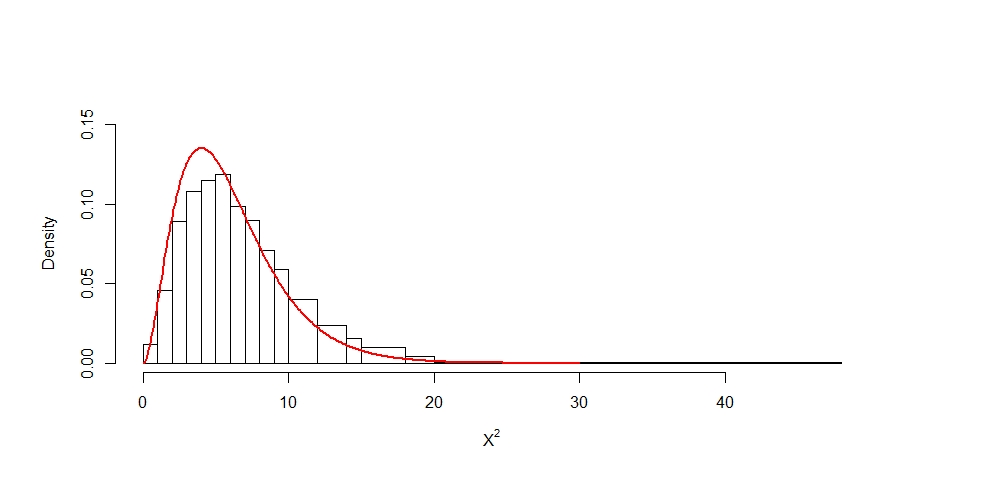}
\caption{The histogram of 10000 $\chi^{2}$ Statistics. The red curve is for $\chi^{2}_{df=6}$}
\label{chi-stat-1000}
\end{figure}

\begin{figure}[H]
	\centering
	\includegraphics[width=1.9in]{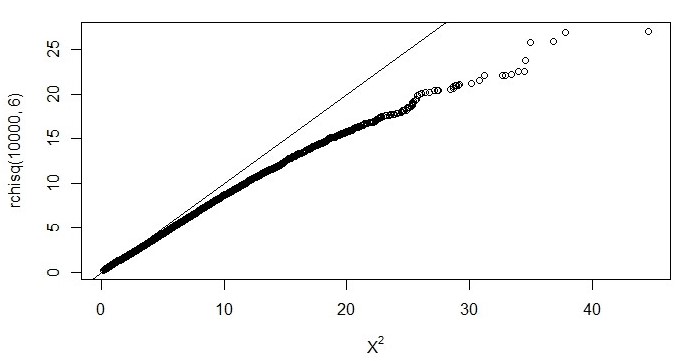}
	\caption{The qq-plot for the 10000 statistics against $\chi^{2}_{df=6}$}
	\label{qq-plot-bin}
\end{figure}

We compare the statistics of 10000 simulated graphs with $\chi^{2}_{df=6}$ at $\alpha=0.05$ level of significance. 9.70\% of them was rejected which is a high percentage comparing with 5\% significant level. That means we have a heavy tailed distribution. Fig. \ref{chi-stat-1000} gives an impression that the two distributions are consistent, however, applying the qq-plot that shown in Fig. \ref{qq-plot-bin} illustrates how far the values departure from an overall linear trend which leads to reject the null hypotheses.   
\setlength{\belowcaptionskip}{-15pt}
\setlength{\belowcaptionskip}{-0.2pt}

\begin{figure}[H]
	\centering
	\includegraphics[width=2.6in]{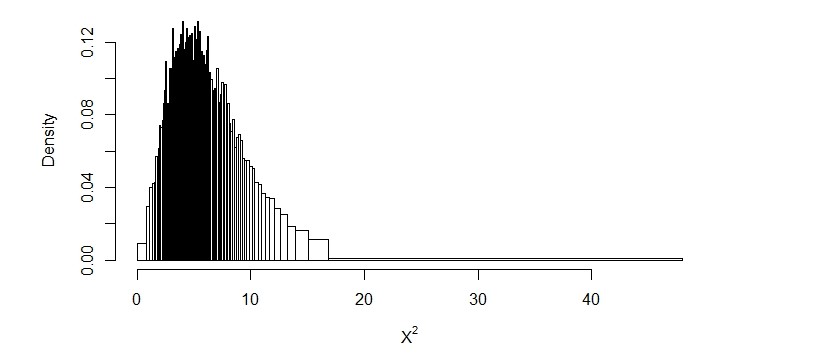}
	\caption{The histogram of 10000 $\chi^2$ scores with 99 quantiles (break points)}
	\label{break}
\end{figure}

We compute the residuals $r_i= (O_i-E_i)^2/E_i$ for 99 quantiles obtaining from $\chi^{2}_{df=6}$ to check how far off are the observed and expected counts \cite{categorical-data}. We use the quantiles' values to break the histogram in Fig. \ref{chi-stat-1000}, and this produces Fig. \ref{break}. We find that there are very large values of residuals that lead to a poor fit. Some of these large contributions concentrate in the first quantiles such as (4.41, 5.29, 4.84, 9.61, 15.21) and the last quantiles such as (21.16, 19.36, 29.16, 17.64, 51.84, 100, 408.04). We conclude that the Binomial dose not fit the distribution of degrees for all nodes collectively in the Erd\H{o}s-Rényi model owing to the fact that the degrees of nodes are not independent.

\section{Correlation of degrees}
\label{corr}
In this section, we calculate the correlation between the degrees of any pair of nodes in the Erd\H{o}s-Rényi graph. Suppose that $X_i$ denotes the degrees of a node $i$, and $X_j$ of a node $j$ such that $i \neq j$. Let $ \delta (i,j)$ represents an indicator function such that $\delta (i,j) = 1$ if $ i, j$ are connected, and $\delta (i,j) = 0 $ if $i, j $ are not connected.

Then $X_i$ can be defined as the total number of connections between $i$ and $j$ other nodes, $ X_i= \sum_{k \neq i} ^{n-1} \delta (i,k)$. Similarly, $ X_j= \sum_{l \neq j} ^{n-1} \delta (j,l)$. The expectation of $X_i$ and $X_j$ are
$E(X_i)=E(X_j)=(n-1)p$ 
and the variances are
$ Var(X_i)=Var(X_j)=(n-1)p (1-p).$ 

To calculate the correlation $\rho_{X_i X_j}$, it is essential to compute first the $Cov(X_i,X_j)$ that is equal to $E(X_i X_j)-E(X_i)E(X_j)$. From the theorem of the expectation of independent random variables, if $X_i$ and $X_j$ are independent, then $E(X_i X_j)=E(X_i)E(X_j)$. However, $X_i$ and $X_j$ are not thoroughly independent, as they perhaps connect to each other. Therefore, two cases of dependency and independency should be considered when calculating $E(X_i X_j)$.

\begin{enumerate}
	\item If $k=l$ 
	\begin{equation*}
	\begin{split}
	E(X_i X_j)
	&= E\left(\sum_{k \neq i} ^{n-1} \delta (i,k) \cdot \sum_{k \neq j} ^{n-1} \delta (j,k) \right) \\
	&= \sum_{k \neq i,j} ^{n-1} E\left(\delta(i,k) \cdot \delta (j,k)\right) \\ 
	&= \sum_{k \neq i,j} ^{n-1} E(\delta(i,k))
	= \sum_{k \neq i,j} ^{n-1} p 
	= p. \\
	\end{split}
	\end{equation*}
	
	Note that $\sum_{k \neq i,j} ^{n-1}=1$, since there is only one case of dependency, that is when $i$ and $j$ are connected.
	
	\item If $k \neq l$ 
	\begin{equation*}
	\begin{split}
	E(X_i X_j)
	&= E\left(\sum_{k \neq i} ^{n-1} \delta (i,k) \cdot \sum_{l \neq j} ^{n-1} \delta (j,l) \right)\\
	&= \sum_{k \neq i} ^{n-1} E(\delta (i,k)) \cdot \sum_{l \neq j} ^{n-1} E( \delta (j,l) ) \\
	&= \sum_{k \neq i} ^{n-1} p \cdot \sum_{l \neq j} ^{n-1} \cdot p 
	= \sum_{k \neq i} ^{n-1} \sum_{l \neq j} ^{n-1} p^2
	= [(n-1)^2 -1]p^2. \\
	\end{split}
	\end{equation*}
\end{enumerate}

In the second case, $E(\delta(i,k) \cdot \delta (j,k))$ can either be $ E(\delta(i,k) \cdot 1)$ or $E(1 \cdot \delta(j,l))$ as $\delta$ is an indicator function. $\sum_{k \neq i} ^{n-1} \sum_{l \neq j} ^{n-1} $ is equal to (the number of dependency cases $1$, subtracting from $(n-1)^2$).
Thus, adding the two cases up, $E(X_i X_j)$ can be written as,
\begin{equation}
E(X_i X_j)= p + [(n-1)^2-1] p^2 
\end{equation}
Consequently,
\begin{equation}
Cov(X_i,X_j)
=E(X_i X_j)-E(X_i)E(X_j)=p (1-p).
\label{COV}
\end{equation}

The correlation between the degrees of any pair of nodes in an Erd\H{o}s-Rényi graph can be obtained as following,
\begin{equation}
\rho_{X_iX_j} =\frac{Cov(X_i,X_j)}{\sqrt {Var(X_i) \times Var(X_j)}}=\frac{1}{(n-1)}.
\label{Corr}
\end{equation} 


\section{Node degree distribution: Testing the multivariate normal MVN as distribution for the whole set of nodes in a graph}
\label{MVN}

It is known that the degree distribution for a single node in the Erd\H{o}s-Rényi model is Bin$(n-1,p)$, and when the number of nodes are large this can be approximated by Normal$\left((n-1)p, (n-1)p(1-p)\right)$.  Therefore, we study the multivariate normal MVN as a distribution for the degrees of all nodes collectively such that 
\begin{equation}
X\sim \mathcal{N}_N(\mu,\Sigma)
\label{dist_mvn}
\end{equation}
where $X=(X_1,X_2,... ,X_N)^T$ is an $N$-dimensional column vector of random variables each of which has a normal approximation to Binomial distribution, and it has $N\times 1$ mean vector such that 
\begin{equation}
\mu=[(n-1)p,...,(n-1)p]^T 
\label{mean_mvn}
\end{equation}
and
$\Sigma$ is the $N\times N$ nonsingular positive definite variance-covariance matrix such that the diagonal of $\Sigma$ is
\begin{equation}
Var(X)= (n-1)p(1-p)
\label{diag_mvn}
\end{equation}
and the off-diagonal elements are
\begin{equation}
 	Cov(X_i,X_j)=
\left\lbrace
\begin{array}{l}
	0, \,\,\,\,\,\,\,\,\,\,\,\,\,\,\,\,\,\, \mbox{if we assume independent degrees} \\
p(1-p),  \mbox{if we assume dependent degrees}
\end{array}
\right.
\label{off_diag_mvn}
 \end{equation}

In our simulation study, we use the following chi square goodness of fit
\begin{equation}
(x-\mu)^T \Sigma ^{-1} (x-\mu) \sim \chi^2_n
\label{z-score-MVN}
\end{equation}
with considering two methods in calculating $\mu$ and $\Sigma$, and each method with two assumptions of degrees: independence and dependence.
\begin{enumerate}
	\item Actual $p$ : we use the actual value of $p$ that has been used to generate graphs.
	\item Estimated $\hat{p}$ : we estimate $\hat{p}$ from each generated graph by taking the ratio between the number of edges and the maximum possible number of edges. In this method we lose one degree of freedom.
\end{enumerate}

\subsection{Testing MVN for $G(61,0.1)$ using actual $p$}
\label{actualp}

\subsubsection*{\bf First assumption: independent degrees}

\begin{figure}[H]
	\centering
	\begin{subfigure}[t]{0.2\textwidth}
		\centering
		\includegraphics[width=\textwidth]{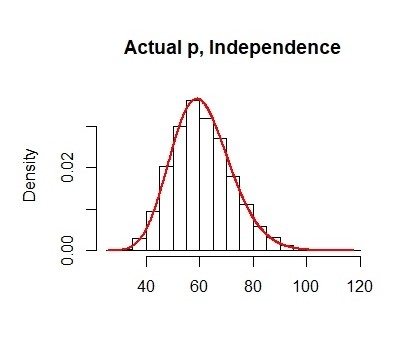}
		\vspace{-2.5em}
		\caption{}
		\label{}
	\end{subfigure}
	\begin{subfigure}[t]{0.2\textwidth}
		\centering
		\includegraphics[width=\textwidth]{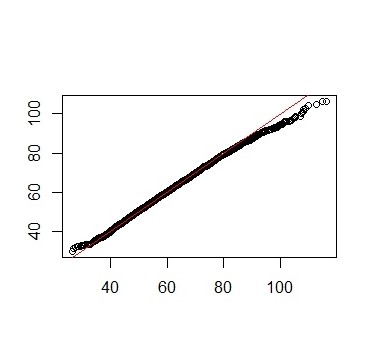}
		\vspace{-2.5em}
		\caption{}
		\label{}
	\end{subfigure} 
	\hspace{1em}
	\caption{A histogram (a) and qq-plot (b) of 10000 scores for  $G(61,0.1)$ against $\chi^2_{df=61,0.05}$ distribution in the case of independence} 
	\label{61Gind}
\end{figure}

In this section, we assume independent degrees and we use actual $p$ to compute Eq.\eqref{z-score-MVN} for 10000 simulated graphs where $n=61$ and $p=0.1$. We find that 604 scores are rejected comparing with the critical value of $\chi^2_{df=61}$ at 5\% level of significance. The histogram in Fig. \ref{61Gind} illustrates the 10000 scores with the red curve of $\chi^2_{df=61,0.05}$. It can be clearly seen that the histogram is consistent with the chi square curve. The corresponding qq-plot shows how the values are highly concentrated on the theoretical distribution,   

\subsubsection*{\bf Second assumption: dependent degrees}
For the same 10000 simulated graphs in the previous section, we repeat the test again but this time we assume dependent degrees. In this case we use $Cov(X_i,X_j)=p(1-p)$ rather than zero. We reject 572 scores and obtain the following Figure, 

\begin{figure}[H]
	\centering
	\begin{subfigure}[t]{0.2\textwidth}
		\centering
		\includegraphics[width=\textwidth]{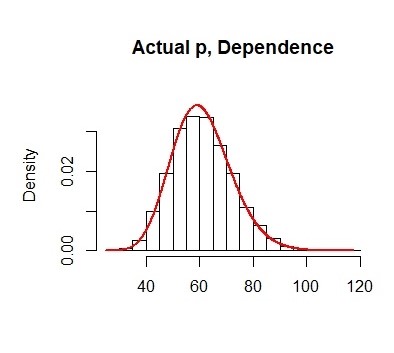}
		\vspace{-2.5em}
		\caption{}
		\label{}
	\end{subfigure}
	\begin{subfigure}[t]{0.2\textwidth}
		\centering
		\includegraphics[width=\textwidth]{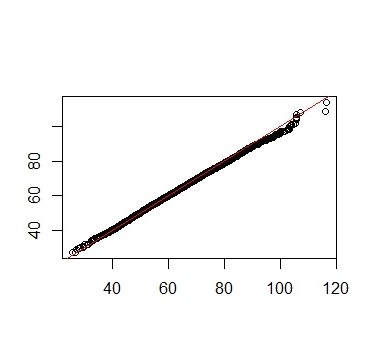}
		\vspace{-2.5em}
		\caption{}
		\label{}
	\end{subfigure} 
	\hspace{1em}
	\caption{A histogram (a) and qq-plot (b) of 10000 scores for  $G(61,0.1)$ against $\chi^2_{df=61,0.05}$ distribution in the case of dependence} 
	\label{}
\end{figure}

\subsection{Testing MVN for $G(61,0.1)$ using estimated $\hat{p}$}
Similarly to Section \ref{actualp} and with the same simulated graphs, we do the test using estimated $\hat{p}$. When we assume independent degrees, 388 scores are rejected, while when we assume dependent degrees, 482 are rejected which is closer to the level of significance 5\%. Both cases show pretty consistence of the scores and the chi square distribution when we plot their histograms and qq-plots.

\subsection{Summary}
The table below summaries the rejection percentages for the same 10000 simulated graphs using two methods (actual $p$ and estimated $\hat{p}$) and each with two assumptions (independent and dependent degrees). The perfect result will be the one that has a rejection percentage closer to the level of significance 5\%. 
All the percentages in the table are around 5\%, and this is an indication of a good fit. Comparing the methods and assumptions we use in terms of the proportion of rejection and their qq-plot, we notice that the case of dependence and estimated $\hat{p}$ gives slightly better fit than the others.

\begin{table}[H]
\centering
\normalsize
	\begin{tabular}{l |l l }
		& Actual $p$  & Estimated $\hat{p}$\\
		\toprule 
		Independence& 6.04\% &3.88\% \\
Dependence& 5.72\% & 4.82\%  \\
		\bottomrule
  \end{tabular}
\caption{The percentages of rejected graphs}
\vspace{-0.79 cm}
\end{table}

\subsection{Testing MVN for various values of $n$ and $p$}

In this section we test different values of $n$ and $p$ using several tests. First, a chi square goodness of fit test in Eq.\eqref{z-score-MVN}, considering 1000 replicates for each specific pairs of $n$ and $p$, and then computing the rejection proportion (Rej. Pro.).

Second, we test 
how close the empirical CDF is to the true CDF of $\chi^2_{df}$ and obtain $p$-values using the Anderson Darling test of goodness of fit
\begin{equation}
A = -n -\frac{1}{n} \sum_{i=1}^{n} [2i-1] [\ln(F_{X_i}) + \ln(1 - F_{n-i+1})]
\end{equation} 
where $F_{i}$ is the cumulative function of the specified distribution, and $i$ is the ith sample calculated when the data is sorted in ascending order \cite{anderson1954test}.

Also, we evaluate the sum of squared difference (Squared Euclidean Distance) between the CDFs, and obtain the scaling values using the following equations,
\begin{equation}
\hat{F_n}(x)=\frac{1}{n}\sum_{i=1}^{n}{\mathds{1}\{ T_i \leq x \}}
\end{equation}
where $\hat{F_n}(x)$ is the empirical distribution function, $\mathds{1} \{A\}$ is the indicator of event $A$, $T_i$ is the $ith$ observed statistic, and $n=1000$.  

\subsubsection*{CDFs similarity}
\begin{equation}
\sum_{i=1}^{n}(\hat{F_n}(x_i)-F(x_i))^2
\end{equation}

\subsubsection*{Scaling}
\begin{equation}
\sum_{i=1}^{n}\frac{(\hat{F_n}(x_i)-F(x_i))^2}{F(x_i)(1-F(x_i))}
\end{equation}

For each pair of $n$ and $p$,
we examine independence and dependence cases, each with actual $p$ and estimated $\hat{p}$.

\subsection*{\bf First case: independent degrees (actual $p$)}

\begin{figure} [H]
	\centerline{\includegraphics[width=7cm]{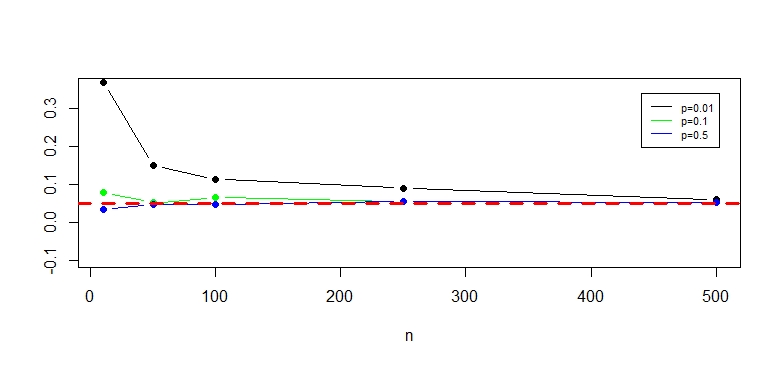}}
	\caption{Proportions of rejected statistics against $n$ in the case of independence with actual $p$, (the red line is the 5\% threshold)}
	\label{Case1}
\end{figure}

\begin{table}[H]
\centering
\normalsize
	\begin{tabular}{l l l  l l l l}
		\toprule
		$p$ & $n$ & $df$& Rej. Pro.& $p$-value & CDFs & Scaling\\
		\midrule 
		&	$10$   & 10     & 0.368    & 6e-07      & 260.9& 5.6e+07\\
		&  $50$    & 50     &  0.149   & 6e-07	    & 11.92  & 3690.7 \\
0.01    &  $100$   & 100    &  0.113   & 6e-07 	    & 4.421   & 64.95   \\  
		&  $250$   & 250	&  0.090   & 1.2e-06 	& 1.145   & 13.54   \\
		&  $500$   & 500	&  0.060   & 4.6e-02	& 0.290   &	2.821    \\
		\midrule  
		&	$10$   & 10	    &  0.077   & 6e-07  & 5.548  & 29.22  \\
		&  $50$	   & 50	    &  0.051   & 0.258 	& 0.197  & 1.478  \\
0.1     &  $100$   & 100	&  0.065   & 0.055 	& 0.340  & 2.356   \\
		&  $250$   & 250	&  0.055   & 0.170 	& 0.316  & 1.606   \\
		&  $500$   & 500	&  0.052   & 0.867  & 0.061  & 0.385   \\
		\midrule  
		&  $10$	   & 10	    &  0.033   & 0.0082 & 1.202 & 7.539 \\
		&  $50$	   & 50	    &  0.046   & 0.6300 & 0.068  & 0.569  \\
0.5     &  $100$   & 100	&  0.047   & 0.9156	& 0.039  & 0.293  \\
		&  $250$   & 250	&  0.054   & 0.1368	& 0.343  & 1.777  \\
		&  $500$   & 500	&  0.053   & 0.9561	& 0.038 & 0.258 \\
		\bottomrule
  	\end{tabular}
		\caption{Independence case with actual $p$}	
		\label{1 case}
\end{table}

\subsection*{\bf Second case: dependent degrees (actual $p$) }

\begin{figure} [H]
	\centerline{\includegraphics[width=7cm]{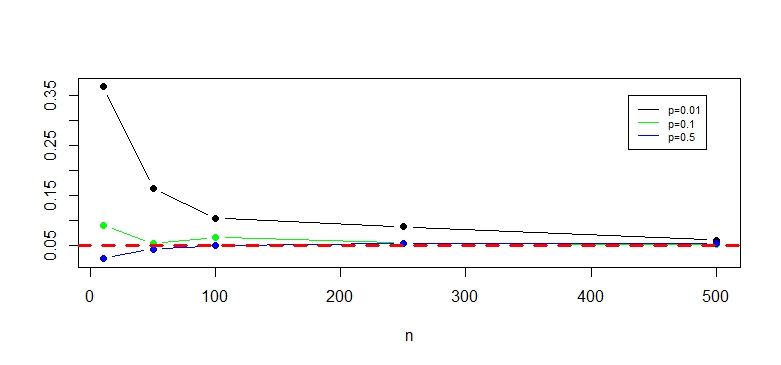}}
	\caption{Proportions of rejected statistics against $n$ in the case of dependence with actual $p$ (the red line is the 5\% threshold)}
	\label{}
\end{figure}

\begin{table}[H]
\centering
\normalsize
			\begin{tabular}{l l l  l l l l}
		\toprule
		$p$ & $n$ & $df$ & Rej. Pro.&$p$-value & CDFs & Scaling\\
		\midrule 
		&  $10$  & 10	 & 0.368  & 6e-07  & 261.9& 3.2e+10\\  
		&  $50$  & 50	 & 0.164  & 6e-07  & 10.83  & 907.8 \\
0.01    &  $100$ & 100   & 0.104  & 6e-07  & 3.992   & 61.46  \\
		&  $250$ & 250   & 0.087  & 7.4e-07  & 1.194   & 13.87  \\ 
		&  $500$ & 500   & 0.060  & 3.8e-02  & 0.312   & 3.015   \\
		\midrule  
		&  $10$  & 10	 & 0.089   & 6e-07    & 4.005  & 24.15 \\ 
		&  $50$  & 50	 & 0.053   & 0.4393    & 0.124  & 0.955  \\
0.1     &  $100$ & 100   & 0.066   & 0.0564    & 0.342  & 2.326  \\
		&  $250$ & 250   & 0.054   & 0.2295    & 0.262  & 1.386  \\
		&  $500$ & 500   & 0.051   & 0.8711	& 0.059  & 0.384  \\
		\midrule  
		&  $10$  & 10	 & 0.024   & 0.0128    & 0.440  & 3.563 \\
		&  $50$  & 50	 & 0.042   & 0.4418  & 0.126   & 0.802 \\
0.5     &  $100$ & 100   & 0.049   & 0.8308  & 0.050   & 0.383 \\
		&  $250$ & 250   & 0.053   & 0.1696  & 0.316   & 1.609 \\
		&  $500$ & 500   & 0.053   & 0.9707  & 0.034   & 0.233 \\
		\bottomrule
  \end{tabular}
			\caption{Dependence case with actual $p$}	
	\label{2 case}
\end{table}

\subsection*{\bf Third case: independent degrees (estimated $\hat{p}$) }

\begin{figure} [H]
	\centerline{\includegraphics[width=7cm]{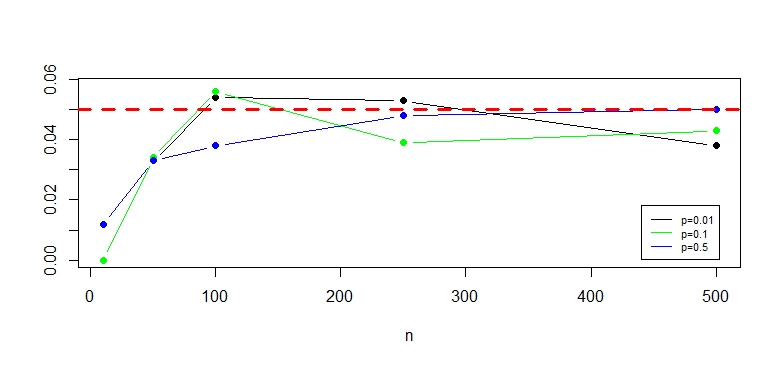}}
	\caption{Proportions of rejected statistics against $n$ in the case of independence with estimated $\hat{p}$ (the red line is the 5\% threshold)}
	\label{}
\end{figure}

\newpage
\begin{table}[htbp]
\centering
\normalsize
	\begin{tabular}{l l l l l l l}
		\toprule
		$p$ & $n$ & $df$ & Rej. Pro.&$p$-value & CDFs & Scaling\\
		\midrule 
		&	$10$   & 9  & 0        & 0.469    & 0.545  & 2.256  \\ 
		&	$50$  & 49  & 0.033  & 1.8e-05 & 1.926 & 9.587  \\
0.01    &  $100$  & 99  & 0.054  & 5.1e-03 & 0.970 & 4.566  \\
		&	$250$ & 249 & 0.053  & 8.4e-01 & 0.062 & 0.460  \\
		&	$500$ & 499 & 0.038  & 6.2e-01 & 0.096 & 0.676  \\
		\midrule  
		&	$10$  & 9   & 0   & 0.8012 & 2.502 & 20.22 \\
		&	$50$  & 49  & 0.034  & 0.0002 & 1.277 & 7.181  \\
0.1     &  $100$  & 99  & 0.056  & 0.8881 & 0.037 & 0.330  \\    
		&	$250$ & 249 & 0.039  & 0.0111 & 0.735 & 3.879  \\
		&	$500$ & 499 & 0.043  & 0.3045 & 0.193 & 1.163  \\
		\midrule  
		&	$10$  & 9   & 0.012  &  6e-07   & 4.957  & 27.25 \\
		&	$50$  & 49  & 0.033  & 0.007  & 0.824  & 4.167 \\
0.5     &	$100$ & 99  & 0.038  & 0.089  & 0.353  & 2.081  \\
		&	$250$ & 249 & 0.048  & 0.006  & 0.906  & 4.567  \\
		&	$500$ & 499 & 0.050  & 0.659  & 0.127  & 0.630  \\
		\bottomrule
  	\end{tabular}
		\caption{Independence case, with estimated $\hat{p}$}
		\label{3 case}	
\end{table}

\subsection*{\bf Fourth case: dependent degrees (estimated $\hat{p}$)}

\begin{figure} [H]
	\centerline{\includegraphics[width=7cm]{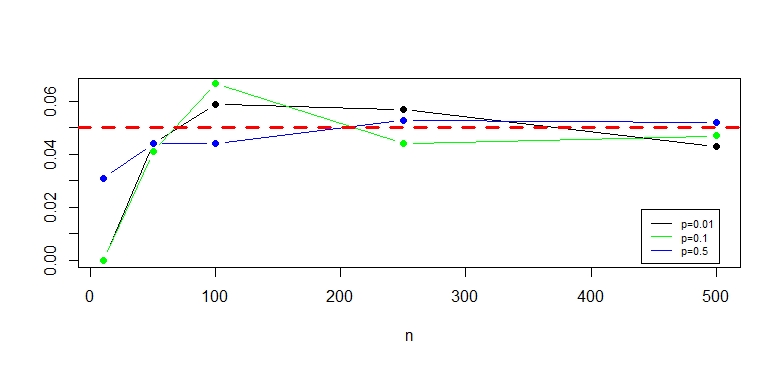}}
	\caption{Proportions of rejected statistics against $n$ in the case of dependence with estimated $\hat{p}$ (the red line is the 5\% threshold)}
	\label{}
\end{figure}

\begin{table}[H]
	\centering
 \normalsize
	\begin{tabular}{l l l l l l l}
		\toprule
		$p$ & $n$ & $df$& Rej. Pro.&$p$-value  & CDFs & Scaling\\
		\midrule 
		&	$10$	& 9	  & 0      & 0.538  &  0.806   & 3.337 \\
		&	$50$	& 49  & 0.044  & 0.004 &	0.685  & 4.312 \\
		0.01    &	$100$	& 99  & 0.059  & 0.288 &	0.172  & 1.162 \\
		&	$250$	& 249 & 0.057  & 0.739 &	0.070  & 0.474 \\
		&	$500$	& 499 & 0.043  & 0.671 &	0.085  & 0.570 \\
		\midrule  
		&	$10$	& 9   & 0      & 0.496 &	2.994 & 35.57 \\
		&	$50$	& 49  & 0.041  & 0.175 &	0.209 & 1.467  \\
		0.1     &	$100$	& 99  & 0.067  & 0.017 &	0.526 & 3.231  \\
		&	$250$	& 249 & 0.044  & 0.295 &	0.214 & 1.181  \\
		&	$500$	& 499 & 0.047  & 0.909 &	0.047 & 0.346  \\
		\midrule  
		&	$10$	& 9   & 0.031  & 0.023 &  0.418 & 2.905 \\
		&	$50$	& 49  & 0.044  & 0.314  &	0.182 & 1.024  \\
		0.5     &	$100$	& 99  & 0.044  & 0.770  &	0.056 & 0.436  \\
		&	$250$	& 249 & 0.053  & 0.120  &	0.293 & 1.490  \\
		&	$500$	& 499 & 0.052  & 0.983  &	0.032 & 0.205  \\
		\bottomrule
	\end{tabular}
	\caption{Dependence case, with estimated $\hat{p}$}
	\label{4 case}	
\end{table}


\section{Results}
Tables \ref{1 case}, \ref{2 case}, \ref{3 case} and \ref{4 case} show numbers of rejected statistics associated with specific values of $n$ and $p$.  We attain the best fit, if 50 out of 1000 statistics are rejected since the level of significance is $5\%$. Any value close to 50 is still acceptable unless it is significantly smaller or larger. Comparing the percentages of the rejected numbers with $5\%$ level of significance, we observe that we always achieve a good fit of multivariate normal distribution with the large values of $n$ and $p$. For example, when $n=500$ and $p=0.5$, the rejected values are 53, 53, 50, 52 for the four cases, respectively. However, we have very poor fit when $n$ or $p$ are very small. Furthermore, the $p$-values of the Anderson Darling test show better fit whenever $n$ and $p$ become larger. Comparing the CDFs, we notice that the two curves in each cases are very similar, and again the quantities improve as $n$ and $p$ increase. Thus, we conclude that multivariate normal is a good approximation for the joint node degree distribution of Erd\H{o}s-Rényi model when $np \geq 10$. Comparing the cases of independent and dependent degrees, most values of Scaling give slightly better results with dependence case.

%
%

\section{Maximum likelihood estimate of an edge presence probability $p$ of MVN for independence and dependence cases}
\label{MLE}

In this section, we compute the maximum likelihood estimate of $p$, the probability of an edge presence in $G(n,p)$, of multivariate normal distribution considering two assumptions: independent and dependent degrees. The only difference in the two cases is the value of $Cov(X_i,X_j)$ in Eq.\eqref{off_diag_mvn}.
If the mean and variance-covariance matrix in Eq.\eqref{dist_mvn} are known, the log likelihood function of an observed vector $X$ is
\begin{equation}
\begin{split}
\log \mathcal{L}(\mu, \Sigma) 
&\propto -\frac{1}{2} \log |\Sigma|-\frac{1}{2} \left((x-\mu)^T\Sigma^{-1}(x-\mu)\right) 
\end{split}
\label{logL}
\end{equation}

First, we have computed $\hat{p}$ from the the moment $\hat{\mu}=\widehat{(n-1)p}=\frac{\sum_{i=1}^{N}x_i}{N}$

\subsubsection*{$\hat{p}$ by moment}

\begin{equation}
\hat p = \frac{\sum_{i=1}^{N}x_i}{N(n-1)}
\label{p_from_mu}
\end{equation}
Next, we find $\hat{p}$ by taking the derivative of Eq.\eqref{logL} with respect to $p$ and equate it to zero in Section \ref{indep_sec} assuming independent degrees and in Section \ref{dep_sec} assuming dependent degrees.

\subsubsection{Independence case}
\label{indep_sec}

In order to compute the maximum likelihood estimate of $p$, we use $\mu$ and $\Sigma$ which are defined in Eqs.\eqref{mean_mvn}, \eqref{diag_mvn}, and \eqref{off_diag_mvn} where we assume independent degrees in this case. First, we need to compute $|\Sigma|$ and $\Sigma^{-1}$ in Eq.\eqref{logL}.
For an $N\times N$ diagonal matrix with all elements in the diagonal equal to a real constant $a$, the determinant of this matrix is $a^N$, and the inverse is $\frac{1}{a}{\bf \text {I}}_{N\times N}$ where ${\bf \text {I}}$ is an identity matrix.
We obtain $|\Sigma|$ and $\Sigma^{-1}$, respectively, as following

\begin{equation}
|\Sigma|=(n-1)^N p^N(1-p)^N
\label{det_sigma_indep}
\end{equation}
\begin{equation}
\Sigma^{-1}=\frac{1}{(n-1) p(1-p)}{\bf \text {I}}_{N\times N}
\label{invers_sigma_indep}
\end{equation}

Then we obtain the following cubic equation after taking the derivative of log likelihood in Eq.\eqref{logL} and equating it to zero
\begin{equation}
\begin{aligned}
&2p^3+\left((n-4)-2\frac{\sum_{i=1}^{N}x_i}{N}\right)p^2
+\left(1+\frac{2\sum_{i=1}^{N}x_i^2}{N(n-1)}\right)p\\
&-\frac{\sum_{i=1}^{N}x_i^2}{N(n-1)}=0 
\label{cube_eq_indep}
\end{aligned}
\end{equation}

To simplify this equation, we can solve it asymptotically ,when $n$ is large, by omitting the cubic term, and all the less significant terms of order $\mathcal{O}(1)$. Then by using $\sqrt{1+y}\approx(1+\frac{y}{2})$ and the assumptions that all the $x_i$ are non-negative and the limitation of the rate of growth of $x_i$ is $\frac{N(n-1)}{2}$, we can obtain the asymptotic answer which is $\frac{\sum_{i=1}^{N}x_i}{N(n-1)}$.


\subsubsection{Dependence case}
\label{dep_sec}

In this section, we obtain $\hat{p}$ assuming dependence between degrees. Therefore, we will use $Cov(X_i,X_j)=p(1-p)$ in $\Sigma$ rather than zero. We need to employ the following results Eqs.\eqref{result3} and \eqref{result4} in order to compute the determinant and inverse of the variance-covariance matrix. 
Let $a,b\in \mathbb{R}$ and $A$ is an $N \times N$ square matrix such that the elements of its diagonal have equal to $a$ and off-diagonal equal to $b$. The determinant and inverse of $A$ are   
\begin{equation}
|A|=[a+(N-1)b](a-b)^{N-1}
\label{result3}
\end{equation}
\begin{equation}
A^{-1}= 
\frac{-b}{(a-b)(Nb+a-b)}{\bf \text {P}}_{N\times N}+\frac{1}{a-b}{\bf \text {I}}_{N\times N}
\label{result4}
\end{equation}
where ${\bf \text {P}}$ is a matrix of ones, i.e. with all entries equal to one, and ${\bf \text {I}}$ is an identity matrix.

Then we obtain
\begin{equation}
\begin{split}
|\Sigma|
&=(n+N-2)(n-2)^{N-1}p^N(1-p)^N\\
\end{split}
\label{det_sigma_dep}
\end{equation}
\begin{equation}
\begin{split}
\Sigma^{-1}
&=\frac{-1}{(n+N-2)(n-2)p(1-p)}{\bf \text {P}}_{N\times N}\\
&+\frac{1}{(n-2)p(1-p)}   {\bf \text {I}}_{N\times N}
\end{split}
\label{invers_sigma_dep}
\end{equation}
where all off-diagonal elements are equal to $-1/\left[(n+N-2)(n-2)p(1-p)\right]$, and diagonal elements are $(n+N-3)/\left[(n+N-2)(n-2)p(1-p)\right]$ 

We differentiate Eq.\eqref{logL} and equate it to zero to get the following cubic equation

\begin{equation}
\begin{aligned}
&4p^3+\left( (n-7)-\frac{2\sum\limits_{i=1}^{N}x_i}{N} \right)p^2 \\
&+2 \left(1+ \frac{(n+N-3)\sum\limits_{i=1}^{N}x_i^2 -\sum\limits_{i=1}^{N}(x_i \sum\limits_{j;j\neq i}^{}x_j)}{N(n-1)(n-2)}  \right)p \\
& + \frac{\sum\limits_{i=1}^{N}(x_i \sum\limits_{j;j\neq i}^{}x_j)-(n+N-3)\sum\limits_{i=1}^{N}x_i^2}{N(n-1)(n-2)} =0
\label{cube_eq_dep}
\end{aligned}
\end{equation}

Similarly to Eq.\eqref{cube_eq_indep}, we can solve Eq.\eqref{cube_eq_dep} asymptotically when $n$ is large. Then we get similar approximation of Eq.\eqref{cube_eq_indep} which is $\approx \frac{\sum_{i=1}^{N}x_i}{N(n-1)}$. In the case of dependence between degrees $N=n$.




\section{Comparing the estimators of the edge presence probability assuming independent degrees Eq.\eqref{cube_eq_indep} and dependent degrees Eq.\eqref{cube_eq_dep}}
\label{Comp}
Due to the fact that the degrees of nodes in a graph are not independent, we expect that the estimator of $p$ assuming dependence in Eq.\eqref{cube_eq_dep} is more accurate than Eq.\eqref{cube_eq_indep} assuming independence. Consecutively, we make a comparison between the two estimators in terms of their bias, variance and mean square error. We generate some Erd\H{o}s-Rényi graphs and use their node degree for $x_i$. We solve the equations numerically using the same dataset $x_i$, $i\in\{1,\cdots,n\}$, and the function {\bf \em uniroot} in R software. We repeat the comparison with various values of $n$ and $p$ particularly when $n$ is very small and very large.  

\subsection{Comparing bias}

\begin{center}
			\begin{figure}[H]
				\centering
			\begin{subfigure}[t]{0.43\textwidth}
				\centering
				\includegraphics[width=\textwidth]{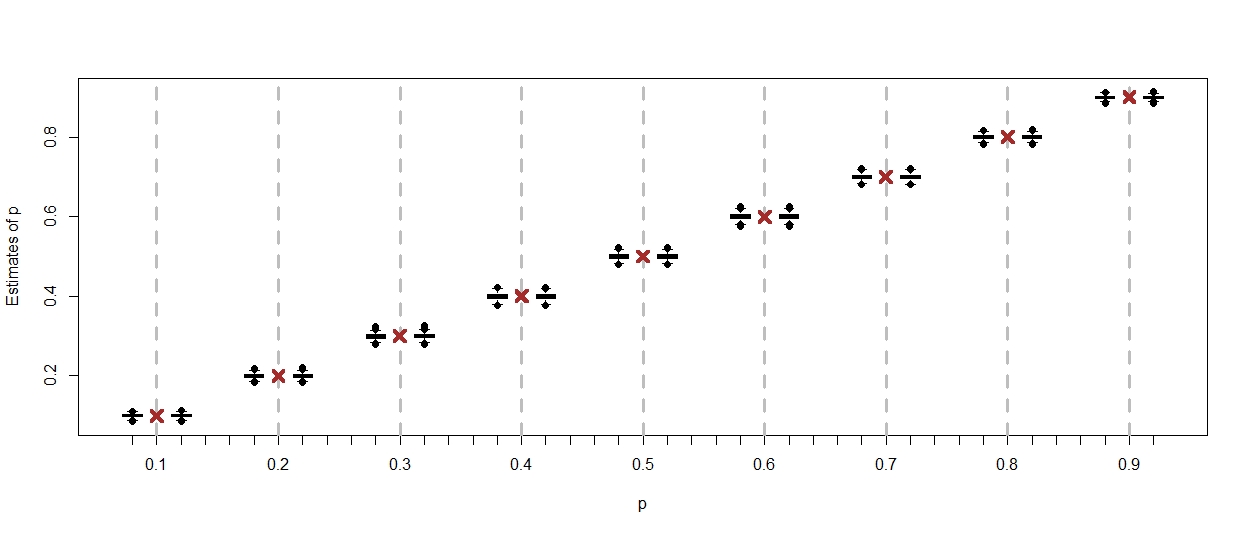}
				\caption{Box-plot of estimates of $p$} 
				\label{large_n_boxplot}
			\end{subfigure}
			\begin{subfigure}[t]{0.29\textwidth}
				\centering
				\includegraphics[width=\textwidth]{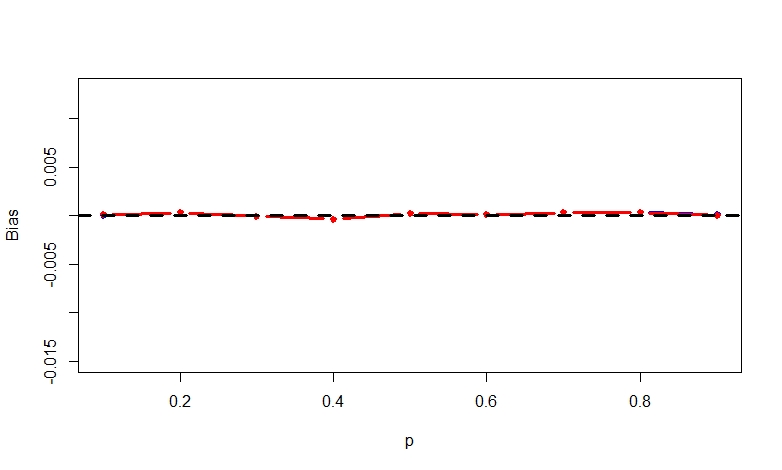}
				\caption{Biases} 
				\label{large_n_bias}
		\end{subfigure} 
		\caption{(a) Compare estimates of $p$ obtained from Eq.\eqref{cube_eq_indep} for independence case (blue), and Eq.\eqref{cube_eq_dep} for dependence case (red) when $n$ is large $n=100$ with various values of $p$. (b) compare their biases}
		\vspace{-3.5em}
		\label{com_bias_n100}
	\end{figure}
\end{center}

When we compare the variance and mean square error of the two estimators, we do not recognize any difference between them. However, when we compare their bias, we observe that Eq.\eqref{cube_eq_dep} gives smaller bias most often principally when $n$ is small. In addition, we have noticed that whatever the value of $n$ is with $p=0.5$, the two estimators have very similar values of bias, variance and mean square error. As $n$ increases, the estimates concentrate around their mean, Fig.\ref{large_n_boxplot}, the difference (smaller bias) starts vanishing and the biases in both cases become closer to zero, Fig.\ref{large_n_bias}. 

\subsection*{\bf Small $n$}
\begin{center}
	\begin{figure}[H]
		\centering
			\begin{subfigure}[t]{0.43\textwidth}
				\centering
				\includegraphics[width=\textwidth]{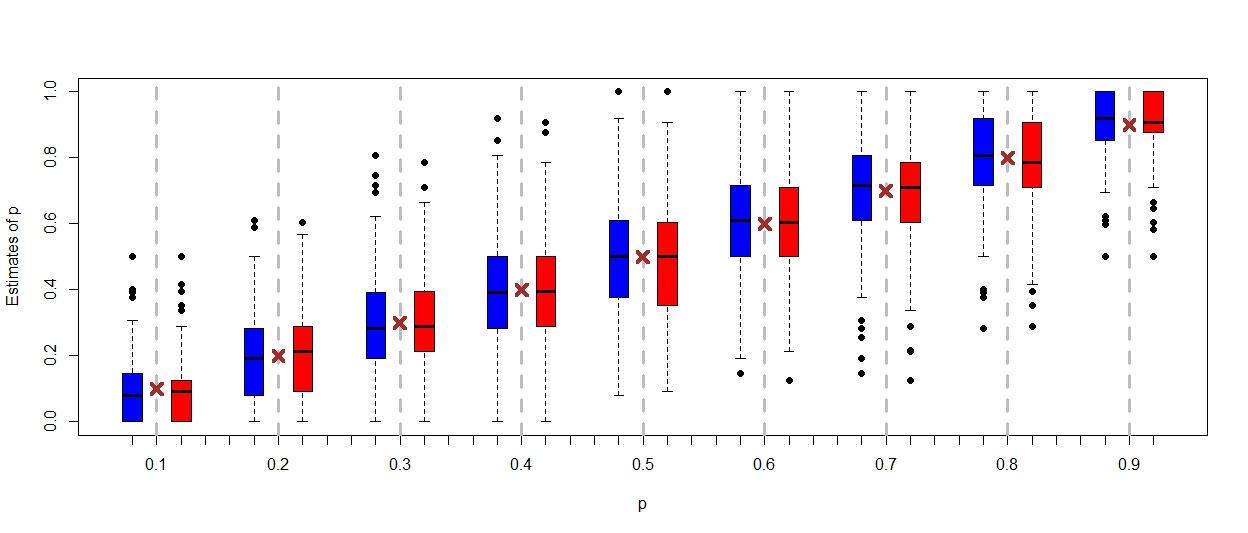}
				\caption{Box-plot of estimates of $p$} 
				\label{small_n_boxplot}
			\end{subfigure}
			\begin{subfigure}[t]{0.29\textwidth}
				\centering
				\includegraphics[width=\textwidth]{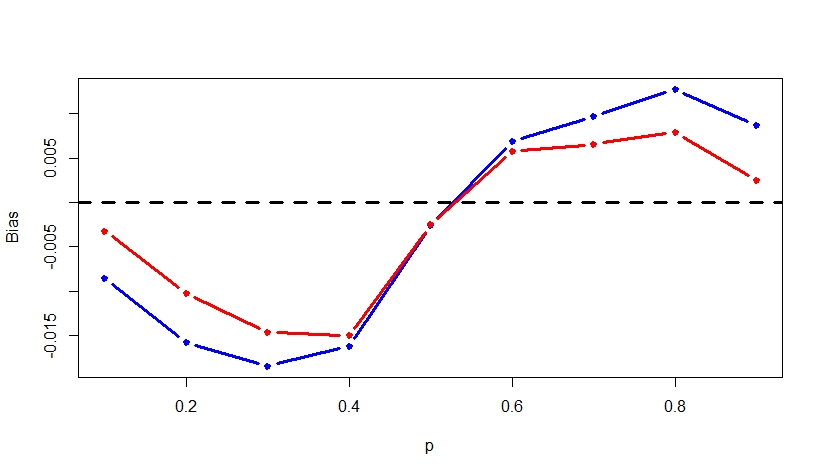}
				\caption{Biases} 
				\label{small_n_gra}
		\end{subfigure} 
				\caption{(a) Compare estimates of $p$ obtained from Eq.\eqref{cube_eq_indep} for independence case (blue), and Eq.\eqref{cube_eq_dep} for dependence case (red) when $n$ is small $n=5$ with various values of $p$. (b) compare their biases}
		\label{com_bias_n5}
	\end{figure}
\end{center}

\subsection*{\bf Large $n$}

\subsection{Paired t-test}

\begin{figure}[H]
	\centering
	\begin{subfigure}[t]{0.23\textwidth}
		\centering
		\includegraphics[width=\textwidth]{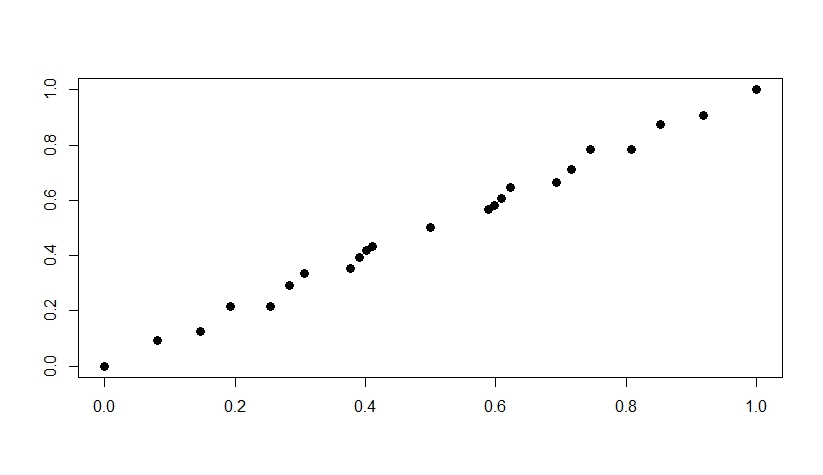}
		\caption{$n=5$}
		\label{}
	\end{subfigure}
	\begin{subfigure}[t]{0.23\textwidth}
		\centering
		\includegraphics[width=\textwidth]{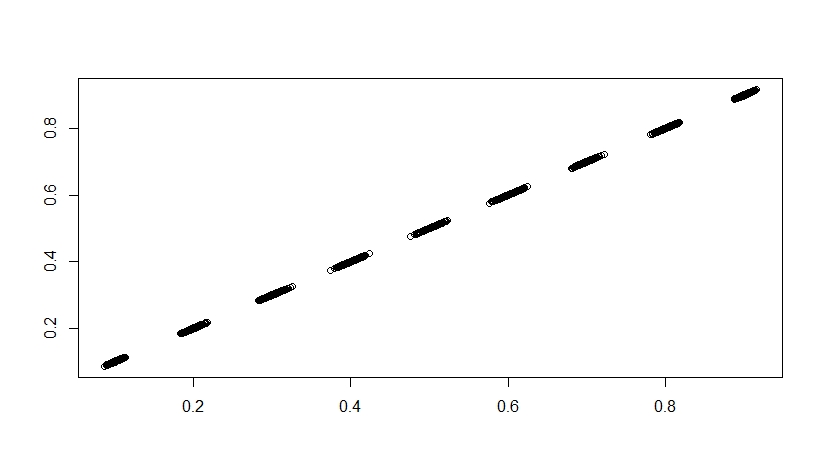}
		\caption{$n=100$}
		\label{}
	\end{subfigure} 
	\caption{Estimates of $p$ of Eq.\eqref{cube_eq_indep} against Eq.\eqref{cube_eq_dep} over 1000 simulated graphs with various values of $p$} 
	\label{scatter}
\end{figure}

For a single dataset, we have an estimate of $p$ using Eq.\eqref{cube_eq_indep} and another estimate of $p$ using Eq.\eqref{cube_eq_dep} which gives one pair. In order to verify the difference between the two estimators, we conduct the paired t-test \cite{david1997paired} and obtain a $p$-value for 10000 pairs. We do the test one time when $n$ is very small $n=5$ and large $n=100$ with changing the value of $p$. In table \ref{p-test}, the $p$-values are significant with small $n$ unless $p$ is close to 0.5. Fig.\ref{scatter} illustrates the scatter plots of the estimates of $p$ using the two estimators with small and large values of $n$. As $n$ decreases, the correlation becomes larger, and the difference between the estimators are more obvious.

\begin{table}[H]
\centering
\small
\begin{tabular}{l|| lllll} 
	\toprule
$p$&0.1&0.3&0.5&0.7&0.9\\
\midrule
$n=5$ & $6.7\times 10^{-6} $&0.143&0.966&0.124&$1.07\times 10^{-5} $ \\
$n=100$& 0.9522 & 0.954& 0.999& 0.991& 0.9451\\
\bottomrule	
\end{tabular}
	\caption{$p$-values of paired t-test to compare the estimates of $p$ obtained from Eq.\eqref{cube_eq_indep} against Eq.\eqref{cube_eq_dep} when $n$ is small and large.}
	\label{p-test}
\end{table}

\section{Conclusion}

We have shown that the degrees of nodes collectively in the Erd\H{o}s-Rényi network are multivariate normal distributed when $np \geq 10$. When $n$ is very small, the maximum likelihood estimator for dependence case Eq.\eqref{cube_eq_dep} is slightly better than the one for independence Eq.\eqref{cube_eq_indep}. However, the difference vanishes when $n$ grows. This is because the dependence between degrees is very weak. 
The correlation equals to $1/(n-1)$ which depends only on $n$ not $p$. As $n$ increases, the correlation decreases and becomes not significant. The highest value of correlation we can gain is when $n$ is very small. 
Knowing the joint node degree distribution can be used to test whether a given graph comes from the Erd\H{o}s-Rényi network. If we have a graph with a specific number of edges,
in that case we can also consider the other version of the Erd\H{o}s-Rényi model which is $G(n,m)$ where $m$ is the exact number of edges.


%





\ifCLASSOPTIONcaptionsoff
  \newpage
\fi




\begin{thebibliography}{10}
\providecommand{\url}[1]{#1}
\csname url@samestyle\endcsname
\providecommand{\newblock}{\relax}
\providecommand{\bibinfo}[2]{#2}
\providecommand{\BIBentrySTDinterwordspacing}{\spaceskip=0pt\relax}
\providecommand{\BIBentryALTinterwordstretchfactor}{4}
\providecommand{\BIBentryALTinterwordspacing}{\spaceskip=\fontdimen2\font plus
\BIBentryALTinterwordstretchfactor\fontdimen3\font minus
  \fontdimen4\font\relax}
\providecommand{\BIBforeignlanguage}[2]{{%
\expandafter\ifx\csname l@#1\endcsname\relax
\typeout{** WARNING: IEEEtran.bst: No hyphenation pattern has been}%
\typeout{** loaded for the language `#1'. Using the pattern for}%
\typeout{** the default language instead.}%
\else
\language=\csname l@#1\endcsname
\fi
#2}}
\providecommand{\BIBdecl}{\relax}
\BIBdecl

\bibitem{ER1}
P.~Erdos and A.~Renyi, ``On random graphs i.'' \emph{Publ. Math. Debrecen},
  vol.~6, pp. 290--297, 1959.

\bibitem{Solomonoff}
R.~Solomonoff and A.~Rapoport, ``Connectivity of random nets,'' \emph{The
  bulletin of mathematical biophysics}, vol.~13, no.~2, pp. 107--117, 1951.

\bibitem{ER2}
P.~Erdos and A.~Renyi, ``On the evolution of random graphs,'' \emph{Publ. Math.
  Inst. Hung. Acad. Sci}, vol.~5, no.~1, pp. 17--60, 1960.

\bibitem{ER3}
------, ``On the strength of connectedness of a random graph,'' \emph{Acta
  Mathematica Hungarica}, vol.~12, no. 1-2, pp. 261--267, 1961.

\bibitem{Networks-Intro-2010}
M.~E.~J. Newman, \emph{\BIBforeignlanguage{English}{Networks: an
  introduction}}.\hskip 1em plus 0.5em minus 0.4em\relax Oxford: Oxford
  University Press, 2010.

\bibitem{fischer2010history}
H.~Fischer, \emph{A history of the central limit theorem: From classical to
  modern probability theory}.\hskip 1em plus 0.5em minus 0.4em\relax Springer
  Science \& Business Media, 2010.

\bibitem{cochran1952chi}
W.~Cochran, ``The chi-square goodness-of-fit test,'' \emph{Annals of
  Mathematical Statistics}, vol.~23, no.~3, pp. 15--345, 1952.

\bibitem{df}
S.~Pandey and C.~L. Bright, ``What are degrees of freedom?'' \emph{Social Work
  Research}, vol.~32, no.~2, pp. 119--128, 2008.

\bibitem{categorical-data}
A.~Agresti, \emph{\BIBforeignlanguage{English}{An introduction to categorical
  data analysis}}, 2nd~ed.\hskip 1em plus 0.5em minus 0.4em\relax Hoboken, NJ:
  Wiley-Interscience, 2007.

\bibitem{anderson1954test}
T.~W. Anderson and D.~A. Darling, ``A test of goodness of fit,'' \emph{Journal
  of the American statistical association}, vol.~49, no. 268, pp. 765--769,
  1954.

\bibitem{david1997paired}
H.~A. David and J.~L. Gunnink, ``The paired t test under artificial pairing,''
  \emph{The American Statistician}, vol.~51, no.~1, pp. 9--12, 1997.

\end{thebibliography}
%


\bibliographystyle{IEEEtran}

\end{document}